\documentclass[10pt]{article}

\usepackage[backend=biber, style=vancouver]{biblatex}
\usepackage{orcidlink}
\usepackage{authblk}

\hypersetup{
    colorlinks=true,
    linkcolor=blue!50!black,
    citecolor=blue!50!black,
    urlcolor=blue!60!black
}

\usepackage[T1]{fontenc}
\usepackage{graphicx,verbatim}

\usepackage{multirow}
\usepackage[table]{xcolor}
\usepackage{nicematrix, tikz}

\usepackage{subcaption}
\usepackage{caption}
\PassOptionsToPackage{hyphens}{url}\usepackage{hyperref}
\usepackage{xurl}

\usepackage{color}

\usepackage{geometry}
\usepackage{booktabs}
\usepackage{makecell} 
\usepackage{xcolor}

\definecolor{cb1}{RGB}{1, 115, 178}
\definecolor{cb2}{RGB}{222, 143, 5}
\definecolor{cb3}{RGB}{2, 158, 115}
\definecolor{cb4}{RGB}{213, 94, 0}
\definecolor{cb5}{RGB}{204, 120, 188}
\definecolor{cb6}{RGB}{202, 145, 97}
\newcommand{\legenddot}[1]{\textcolor{#1}{\textbullet}}

\newif\ifPrintFigures
\PrintFigurestrue   

\newif\ifPrintTables
\PrintTablestrue

\begin{document}
\title{Benchmarking Foundation Models for Renal Lesion Stratification in CT}

\author{Hartmut Häntze\textsuperscript{1,2,3}~\orcidlink{0000-0002-6204-571X}}
\author{Sarah de Boer\textsuperscript{2}~\orcidlink{0000-0001-5184-4340}}
\author{Myrthe Buser\textsuperscript{2}~\orcidlink{0000-0003-0640-6434}}
\author{Alessa Hering\textsuperscript{2}~\orcidlink{0000-0002-7602-803X}}
\author{Bram van Ginneken\textsuperscript{2}~\orcidlink{0000-0003-2028-8972}}
\author{Mathias Prokop\textsuperscript{2}~\orcidlink{0000-0001-8157-8055}}
\author{Jawed Nawabi\textsuperscript{4}~\orcidlink{0000-0002-1137-0643}}
\author{Sebastian Ziegelmayer\textsuperscript{3}~\orcidlink{0000-0001-8724-4718}}
\author{Lisa Adams\textsuperscript{3}~\orcidlink{0000-0001-5836-4542}}
\author{Keno Bressem\textsuperscript{3,5}~\orcidlink{0000-0001-9249-8624}}

\affil{%
\textsuperscript{1}Charité - Universitätsmedizin Berlin, Department of Radiology, Berlin, Germany \\
\textsuperscript{2}Radboudumc, Diagnostic Image Analysis Group, Nijmegen, The Netherlands \\
\textsuperscript{3}Klinikum rechts der Isar, TUM University Hospital, Technical University of Munich, Munich, Germany \\
\textsuperscript{4}Charité - Universitätsmedizin Berlin, Institute of Neuroradiology, Berlin, Germany \\
\textsuperscript{5}German Heart Center, TUM University Hospital, Technical University of Munich, Munich, Germany%
}

\date{Corresponding author: hartmut.haentze@charite.de}

\maketitle              

\begin{abstract}
\normalsize
The rapid proliferation of open-source medical foundation models (FMs) raises a practical question: how well do their pre-trained representations transfer to clinically relevant but data-scarce classification tasks? Particularly in CT-based renal lesion classification, a push toward greater generalizability would be meaningful, as the field is constrained by inherently limited training data. We addressed this through a benchmark of three medical FMs on this specific task. This six-class problem spans common entities like cysts and clear cell renal cell carcinoma, alongside rare subtypes. Using a frozen feature-probing protocol, we compared FM embeddings against a handcrafted radiomics classifier and a 3D ResNet-50 trained from scratch. Models were trained on a composite dataset of 2,854 lesions and evaluated on an external test set of 234 lesions from The Cancer Imaging Archive. Our results reveal two key findings. First, FM performance (AUC 0.70–0.77) matched the from-scratch ResNet (AUC 0.72) while drastically reducing hardware demand, requiring only seconds on a CPU after feature extraction. However, the conventional radiomics baseline significantly outperformed all deep learning approaches, achieving an AUC of 0.88 (all p $\leq$ 0.002). This suggests that current generalist FM embeddings do not yet capture the fine-grained texture and shape heterogeneity driving histological subtype discrimination. Despite their potential in data-scarce settings, medical FMs did not surpass established models for renal lesion stratification, leaving radiomics as the current state-of-the-art.
\end{abstract}

\begin{center}
\end{center}

\section{Introduction}
Medical foundation models (FMs) present a new methodological alternative in automated diagnosis. By leveraging large-scale pre-training, these models aim to extract robust feature representations from datasets that are otherwise too small for training deep neural networks from scratch \supercite{lipkova2024age,schafer2024overcoming}.
Renal lesion stratification offers a clinically relevant test case for this approach. Kidney cancer ranks as the 14th most frequent malignancy globally and causes the 16th highest number of cancer-related deaths \supercite{iarc_kidney_fact_sheet}. Malignant renal lesions are frequently discovered incidentally, yet their non-invasive characterization remains clinically challenging. While guidelines recommend contrast-enhanced CT \supercite{wang2018ct,powles2024renal,s3_renal_cell_carcinoma_2024}, and radiologists can reliably distinguish benign cysts and neoplasms such as angiomyolipoma from other tumours \supercite{silverman2019bosniak,jinzaki2014renal}, stratifying specific lesion subtypes is often difficult. For instance, differentiating papillary renal cell carcinoma (pRCC) Type I from other pRCC entities \supercite{egbert2013differentiation}, or distinguishing benign renal oncocytoma (RO) from malignant chromophobe RCC (chrRCC), is often not feasible based solely on CT \supercite{choudhary2009renal} or MRI \supercite{rosenkrantz2010mri} features. Consequently, confirming the lesion subtype often necessitates biopsy or surgical resection. In frail or elderly populations, the procedural risks may outweigh the benefits, making a non-invasive, image-based active surveillance strategy highly desirable.

Deep learning approaches have demonstrated potential in addressing these diagnostic gaps. Existing methods generally fall into two categories: end-to-end models predicting cancer type directly from images \supercite{uhm2021deep}, or radiomics-based approaches using engineered features \supercite{li2020value}. While some models target specific binary classifications, such as RO versus chRCC \supercite{li2020value,alhussaini2022comparative,uchida2022apparent} or pRCC type I versus type II \supercite{doshi2016use,gao2022differential}, others attempt multiclass stratification \supercite{uhm2021deep}. These models have notably achieved higher accuracy than radiologists in specific tasks \supercite{uhm2021deep}. However, their broader clinical translation is hindered by the low prevalence of specific histological subtypes. Dataset sizes for rare conditions like RO are often negligible, ranging from 45 cases \supercite{uhm2021deep} to as few as 17 cases in cross-validation settings \supercite{li2020value}. This data scarcity limits the performance of standard convolutional neural networks trained from scratch. For example, Uhm et al. \supercite{uhm2021deep} reported an AUC of 0.77 for chrRCC. While superior to human readers, this metric remains insufficient for reliable automated decision support.

Addressing these performance limitations requires training methodologies capable of learning robust representations from such limited data. In this study, we therefore investigate the efficacy of three open-source FMs for renal lesion stratification primarily using a frozen feature-probing protocol, in line with FM benchmarking practice \supercite{stegeman2026designing}. We benchmark these models against two established baselines: a handcrafted radiomics classifier and a 3D ResNet-50 neural network trained from scratch. All models are implemented within our open-source framework, RenalVision, available at \url{https://github.com/hhaentze/RenalVision}.

\section{Materials and Methods}

\subsection{Study Design and Participants}
This retrospective study was approved by the local ethics committee (EA4/062/20) and conducted in accordance with the Declaration of Helsinki. Patient consent was waived due to the retrospective nature of the study. 

Our training dataset was compiled from two sources. First, we used an in-house cohort of patients with renal masses who underwent CT between 2007 and 2022. We selected arterial and delayed phases and excluded patients with ambiguous histology, multiple solid tumour types, or lesions $<1$ cm in diameter, as these are typically uncharacterisable on CT \supercite{silverman2015incompletely}. tumours and ipsilateral cysts were delineated by a student reader, with tumour subtypes derived from histopathology reports. Second, we incorporated the KiTS23 dataset \supercite{kits23_challenge_2023} (late arterial/nephrogenic phase CTs) applying the same $<1$ cm exclusion criterion. The combined training dataset comprised 2,854 lesions from 1,279 participants.

\begin{figure}[htbp]
\centering
\ifPrintFigures
    \includegraphics[width=\textwidth]{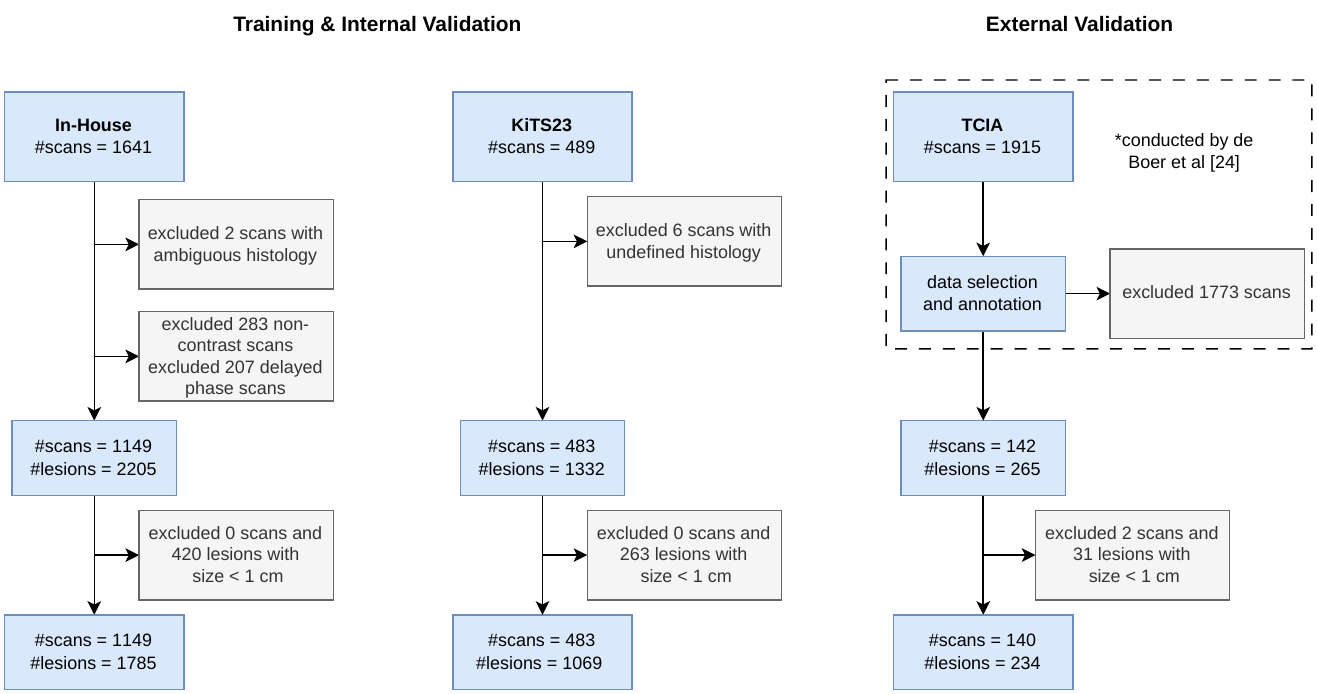}
    \caption{Inclusion and exclusion criteria for the different datasets.}
\else
    \refstepcounter{figure}
\fi
\label{fig:dataflow}
\end{figure}

For external evaluation, we utilized a publicly available dataset of kidney lesions from The Cancer Imaging Archive (TCIA) \supercite{akin2016cancer,linehan2016cancer,linehan2016cancer2} processed by de Boer et al. \supercite{deBoer2026},  who paired automated segmentation \supercite{de2025robust} with human readers to annotate renal lesions. We applied our $<1$ cm exclusion criterion to yield an external test set of 234 lesions across 140 CT scans from 101 participants. The exact dataset compositions are detailed in Table \ref{tab:demographics} and the exclusion criteria in Figure \ref{fig:dataflow}.

\subsection{Lesion Subtype Selection}
Lesions were categorized into six classes: cysts, clear cell (ccRCC), papillary (pRCC), chromophobe (chrRCC), renal oncocytoma (RO), and "Other". Angiomyolipoma, duct tumours, and rare subtypes were merged into the "Other" class due to insufficient representation. The external evaluation data from TCIA is limited to cysts and the three RCC subtypes.

\subsection{Model Selection and Feature Selection}

\begin{figure}[htbp]
\centering
\ifPrintFigures
    \includegraphics[width=\textwidth]{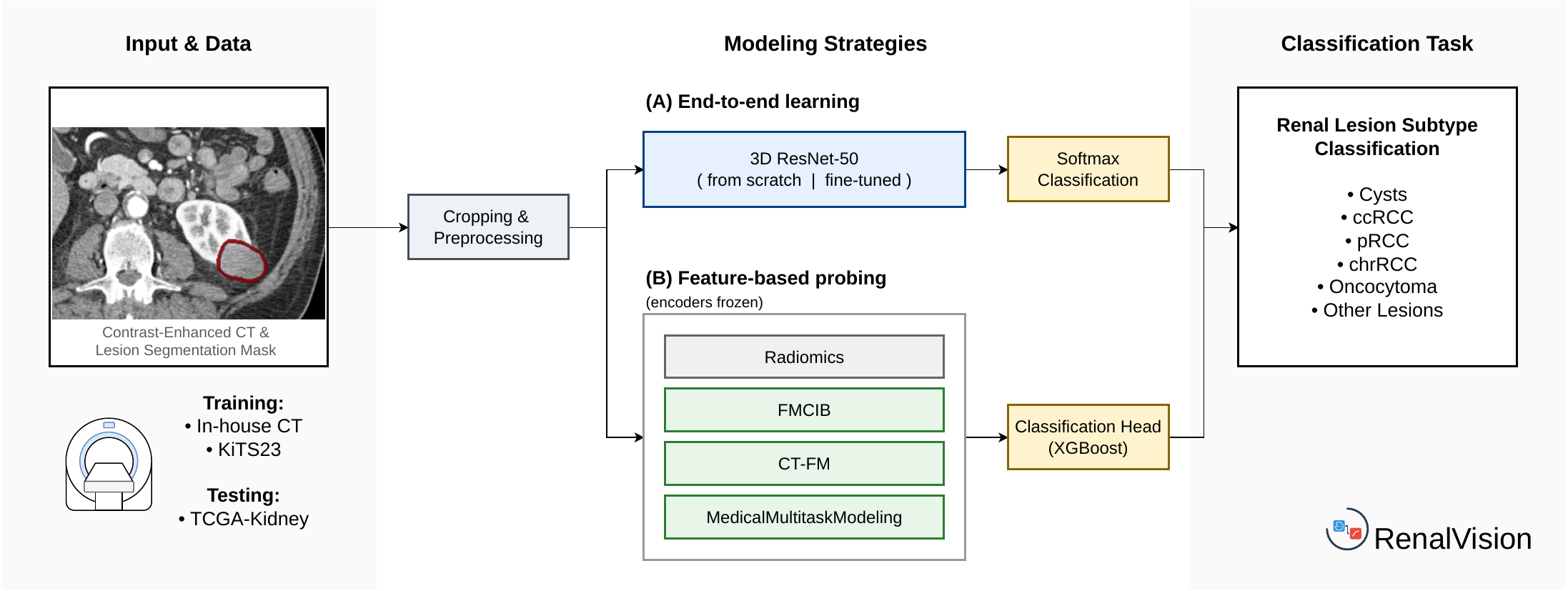}
    \caption{Overview of the processing pipeline. Input data consists of contrast enhanced CT scans with corresponding lesion masks. All pipelines begin with cropping and preprocessing tailored to the specific model architecture. Two modelling strategies are evaluated: (A) End-to-end learning, utilizing a 3D ResNet-50 trained directly on image data; and (B) Feature-based probing, where embeddings are extracted via four distinct encoders and passed to a classification head. Target classes include cysts, clear cell Renal Cell Carcinoma (ccRCC), papillary RCC (pRCC), chromophobe RCC (chrRCC), oncocytoma (RO), and a shared group "Other" for remaining lesion types.}
\else
    \refstepcounter{figure}
\fi
\label{fig:workflow}
\end{figure}

We selected three FMs with open-source weights available at the time of study design:
\textbf{FMCIB} (Foundation Model for Cancer Imaging Biomarkers) \supercite{pai2024foundation},
\textbf{CT-FM} (Vision Foundation Models for Computed Tomography) \supercite{pai2025vision}, and
\textbf{MMM} (MedicalMultitaskModeling) \supercite{schafer2024overcoming,nicke2024tissue}. Two modelling strategies were evaluated that share a common preprocessing pipeline. Images were cropped to the lesion bounding box with a $1$ cm margin , expanded to a minimum $50 \times 50 \times 50$ voxels for small lesions. Random augmentations were applied consistently during training (Strategy A) and feature extraction (Strategy B). Strategy B served as the primary configuration for assessing FMs. Because FMCIB provided replicable open-source implementation code, it was additionally selected for fine-tuning (Strategy A) in accordance with the authors' recommendations. 
The overall workflow is illustrated in Figure \ref{fig:workflow}.

\subsubsection*{Strategy A: End-to-End Learning}
We utilized a 3D ResNet-50 backbone under two initialization schemes.
The \textbf{ResNet Baseline} used inflated ImageNet weights with  a compact classification head (hidden dims: $[2048, 256, 6]$) and an initial learning rate of 1e-4. To assess medical pre-training, the \textbf{FMCIB-fine-tuned} model used FMCIB weights with a larger head (hidden dims: $[4096, 2048, 6]$, by the authors' recommendations) and a split learning rate (1e-6 backbone, 1e-4 head) to preserve pre-trained features. Both schemes addressed class imbalance via weighted sampling and a weighted loss function.

\subsubsection*{Strategy B: Frozen Feature Extraction}
FMs and a radiomics pipeline served as fixed feature encoders for XGBoost \supercite{chen2016xgboost} classifiers utilizing class-balanced loss weighting.
To ensure fair comparisons across varying embedding dimensions (107--4096), hyperparameters were optimized via 10-fold cross-validation.
Preprocessing was adapted to each model's native requirements:
\textbf{FMCIB:} Input volumes were resized to $50 \times 50 \times 50$ voxels, from which the model extracted 4096 features.
\textbf{CT-FM:} The model extracted 512 features utilizing its native sliding-window and embedding-aggregation logic.
\textbf{MMM:} Inputs were resized to $64 \times 64 \times Z$, where Z represents the number of 1 mm axial slices. The model extracted 512 features per three slices, which we aggregated via max pooling and a lesion area–weighted average into a 1024-dimensional embedding.
\textbf{Radiomics:} Lesion volumes were resampled to 1 mm isotropic spacing and clipped to a soft-tissue window ([-160, 240] HU). We extracted 107 distinct features using the default preset of PyRadiomics \supercite{van2017computational}.

\subsection{Statistical Analysis}
Performance was evaluated using class-wise and macro-averaged Area Under the Receiver Operating Characteristic curve (AUC) and Average Precision (AP), alongside a majority-voting baseline. Internal validity was assessed via 10-fold cross-validation on the training set, with 95\% confidence intervals (CIs) derived from Student’s t-distribution. As these folds also guided hyperparameter tuning, internal metrics may be optimistically biased; thus, the held-out external test set serves as our primary measure of generalization. 

External validation involved retraining models on the full training corpus and evaluating them on the held-out TCIA set. Non-parametric bootstrapping (10,000 resamples) \supercite{carpenter2000bootstrap} was used to estimate 95\% CIs. Differences in macro-average AUCs between FMs and baselines were assessed for statistical superiority using simultaneous 95\% CIs via a max-type percentile bootstrap to control the family-wise error rate \supercite{mandel2008simultaneous}. Feature separability was qualitatively assessed using Uniform Manifold Approximation and Projection (UMAP) \supercite{mcinnes2018umap} following Principal Component Analysis. 

\section{Results}
\subsection{Study Population and Demographics}
A total of 1,379 patients were included in this study (Table \ref{tab:demographics}). The combined training cohort comprised 1,279 patients, while the external validation cohort consisted of 100 patients. In the in-house dataset, the mean patient age was 63.6 ± 11.4 years, comprising 548 men and 248 women. The KiTS23 training cohort had a mean age of 59.2 ± 14.4 years (300 men, 182 women), and the TCIA test cohort had a mean age of 55.9 ± 12.7 years (74 men, 26 women). Detailed documentation of the dataset curation is provided in the study flow diagram (Figure \ref{fig:dataflow}).

\begin{table}[htbp]
\centering
\ifPrintTables
    \caption{Patient Demographics and Dataset Characteristics across training and testing cohorts. Values are presented as mean $\pm$ standard deviation for continuous variables and n (\%) for categorical variables.}
    \label{tab:demographics}
    \small
    \setlength{\tabcolsep}{10pt}
    \begin{NiceTabular}{lccc}[colortbl-like]
    \CodeBefore
      \rowcolor{gray!20}{1} 
    \Body
    \hline
    \textbf{Characteristic} & \textbf{In-house} & \textbf{KiTS23} & \textbf{TCIA} \\
    \hline
    \textbf{Patients, n} & 796 & 483 & 100 \\
    \textbf{Age (years)} & 63.6 $\pm$ 11.4& 59.2 $\pm$ 14.4&  55.9 $\pm$ 12.7\\
    \textbf{Sex, n (\%)} & & & \\
    \quad Men & 548 (68.8\%) & 300* (62\%)& 74 (74\%)\\
    \quad Women & 248 (31.2\%) & 182* (38\%)& 26 (26\%)\\
    \textbf{Total Scans, n} &  1149& 483& 140\\
    \textbf{Total Lesions, n} & 1785 & 1069 & 234 \\
    \textbf{Lesion Subtypes, n (\%)} & & & \\
    \quad Cysts & 607 (34.0\%) & 525 (49.1\%) & 94 (40.2\%) \\
    \quad ccRCC & 824 (46.2\%) & 349 (32.6\%) & 94 (40.2\%) \\
    \quad pRCC & 174 (9.7\%) & 51 (4.8\%) & 28 (12.0\%)  \\
    \quad chrRCC & 57 (3.2\%) & 39 (3.6\%) & 18 (7.7\%)  \\
    \quad RO & 84 (4.7\%) & 28 (2.6\%) & 0 (0\%) \\
    \quad Other**& 39 (2.2\%) & 77 (7.2\%) & 0 (0\%) \\
    \textbf{Usage} & internal & internal & external \\
    \hline
    \end{NiceTabular}
    \vspace{2pt}
    \begin{flushleft}
    \footnotesize *One patient in KiTS23 is recorded as transgender (male-to-female) and is excluded from the binary sex stratification, but included in the total cohort count. 
    **Other includes angiomyolipoma, duct tumours, and rare subtypes excluded from primary targets.
    \end{flushleft}
\else
    \refstepcounter{table}
    \label{tab:demographics}
\fi
\end{table}

\subsection{Overall Performance}
We evaluated model performance using 10-fold cross-validation on the internal training set (6 classes) and on the external TCIA test set (4 classes: Cysts, ccRCC, pRCC, chrRCC). Table \ref{tab:combined_results} summarizes the internal validation results. The radiomics baseline achieved the highest performance with a mean AUC of 0.84 and mean AP of 0.45. Among the FMs, MMM yielded the best results (AUC 0.75), outperforming the trained-from-scratch ResNet baseline (AUC 0.64). However, neither FM (probing or fine-tuning) surpassed the radiomics baseline. Notably, initialising the ResNet with FMCIB weights compared with end-to-end fine-tuning resulted in better performance compare to training from scratch (AUC 0.68 vs. 0.64), however, could not match frozen probing (AUC 0.71).

External testing on the TCIA dataset (Table \ref{tab:combined_results}) confirmed the stability of these rankings. Note that the absence of RO and "Others" leads to higher overall macro averages. To contextualise: excluding these classes from the validation data would increase the metrics of the radiomics model from an AUC of 0.84 (AP 0.45) to an AUC of 0.89 (AP 0.63). Despite this baseline shift, the radiomics model remained the top performer on the external set (AUC 0.88, AP 0.64), aligning with the 4-class validation performance. It significantly outperformed the best FM (MMM, AUC 0.77) with a mean difference of 0.11 (95\% CI [0.03, 0.18], Table \ref{tab:sig}), as well as the other FMs. On the lower end, all FMs performed similarly to the ResNet baseline (AUC 0.72). The highest-ranked FM achieved an AUC higher by 0.05, though this did not reach statistical significance (95\% CI [-0.13, 0.02]). 

\subsection{Class-wise Analysis \& Feature Importance}
\begin{figure}[htbp]
\ifPrintFigures
    \centering
    \begin{subfigure}[b]{0.22\linewidth}
        \includegraphics[width=\linewidth]{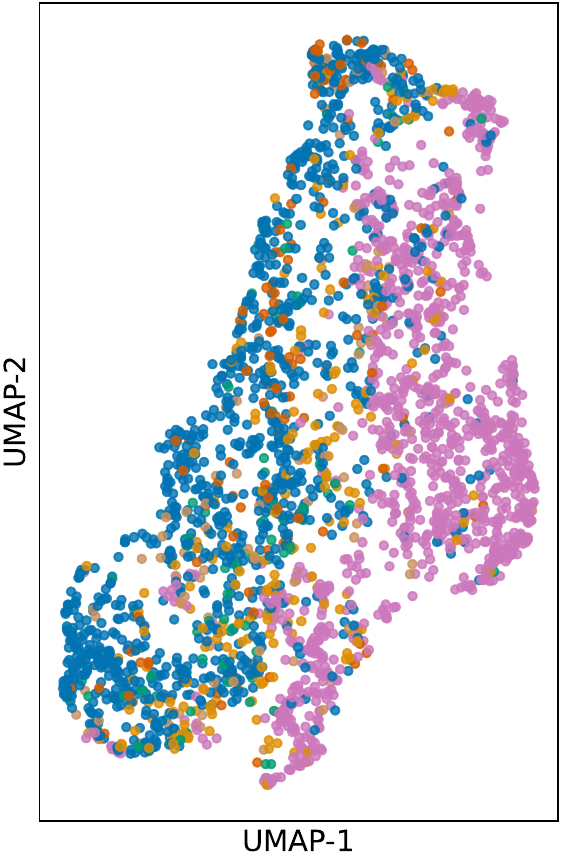}
        \caption{Radiomics}
    \end{subfigure}
    \hfill
    \begin{subfigure}[b]{0.22\linewidth}
        \includegraphics[width=\linewidth]{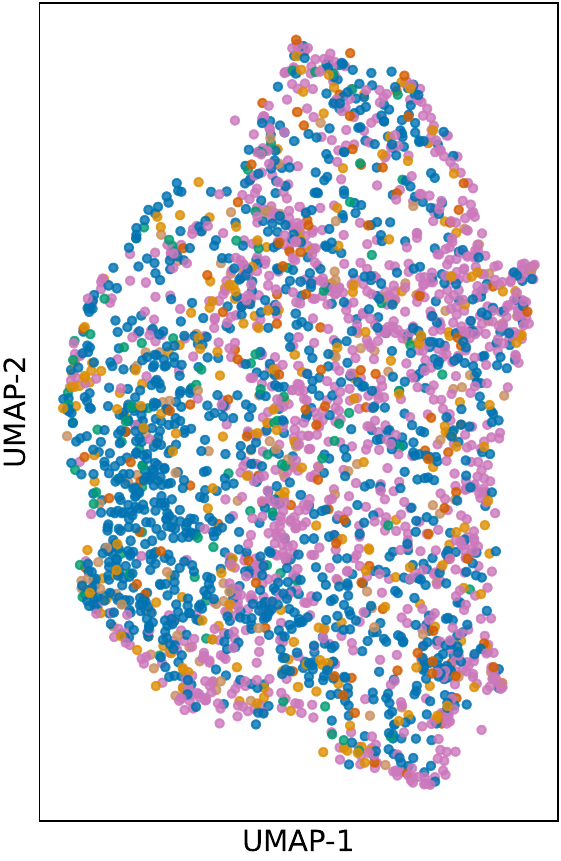}
        \caption{MMM}
    \end{subfigure}
    \hfill
    \begin{subfigure}[b]{0.22\linewidth}
        \includegraphics[width=\linewidth]{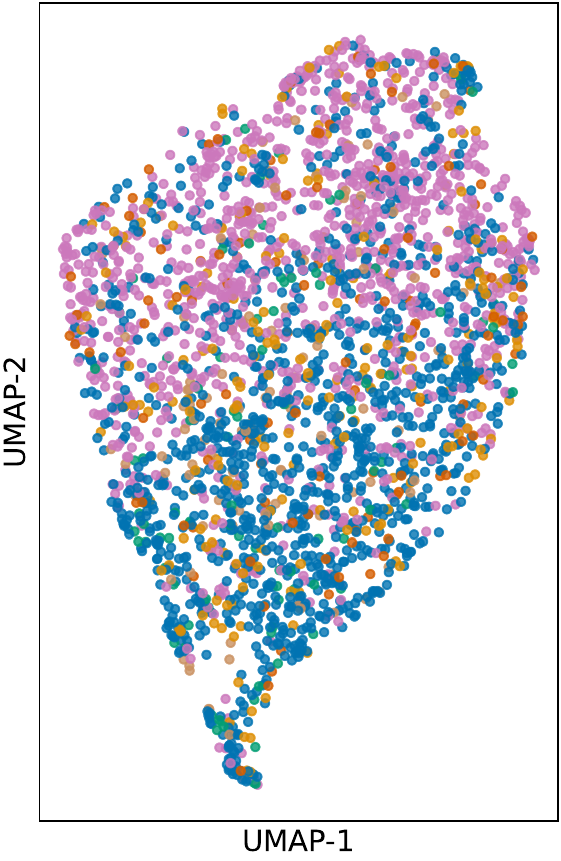}
        \caption{FMCIB}
    \end{subfigure}
    \hfill
    \begin{subfigure}[b]{0.22\linewidth}
        \includegraphics[width=\linewidth]{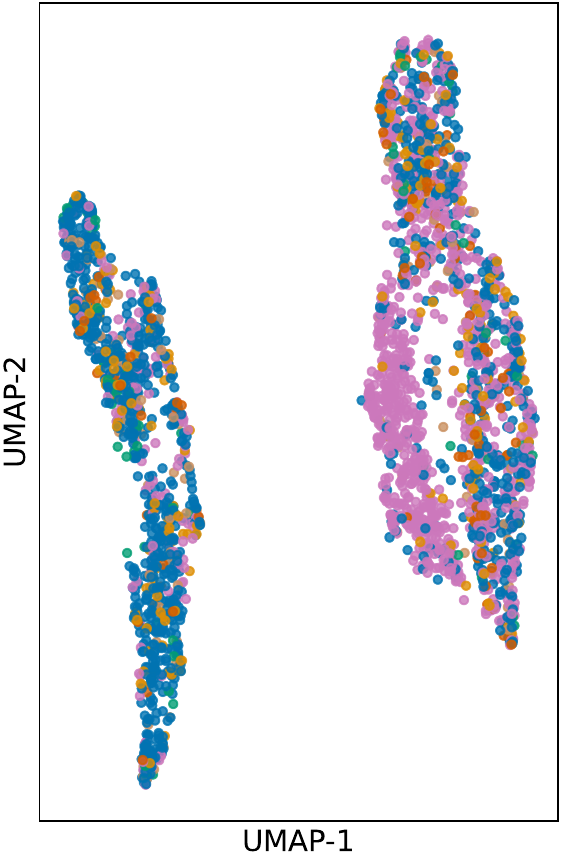}
        \caption{CTFM}
    \end{subfigure}    
    \caption{Low-dimensional UMAP embeddings comparing (a) Radiomics, (b) MMM, (c) FMCIB, and (d) CTFM features. Distinct clusters indicate strong class separability, while overlapping regions suggest feature similarity. Classes are: \legenddot{cb5} Cysts, \legenddot{cb1} ccRCC, \legenddot{cb2} pRCC, \legenddot{cb3} chrRCC, \legenddot{cb4} RO, and \legenddot{cb6} Others. }
\else
    \refstepcounter{figure}
\fi
\label{fig:UMAP_embeddings}
\end{figure}

\begin{figure}[htbp]
\centering
\ifPrintFigures
    \includegraphics[width=0.48\textwidth]{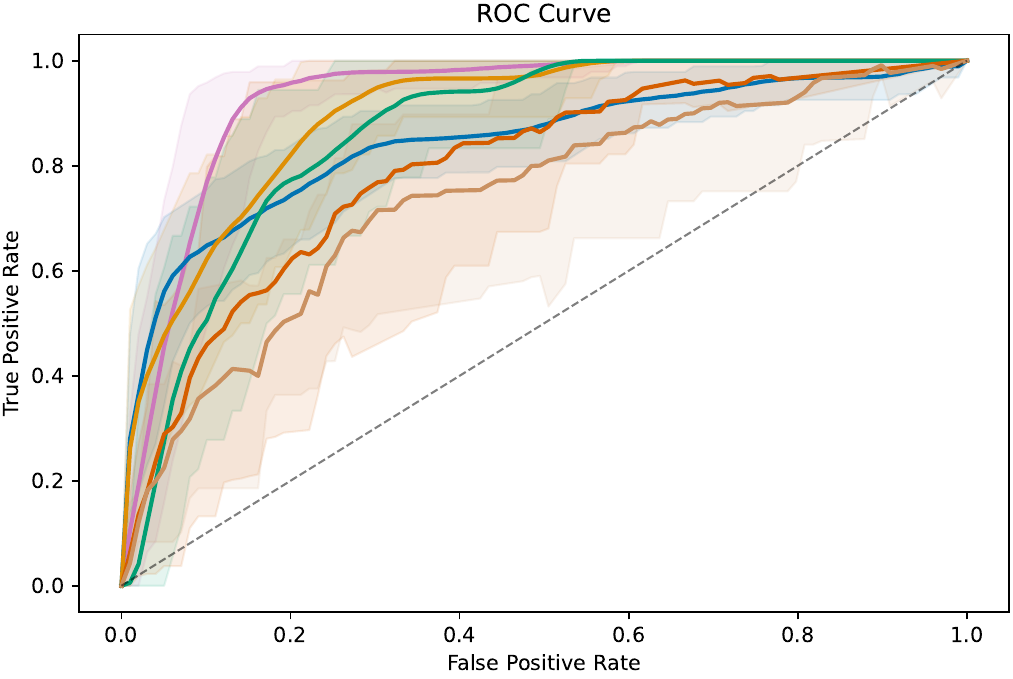}
    \includegraphics[width=0.48\textwidth]{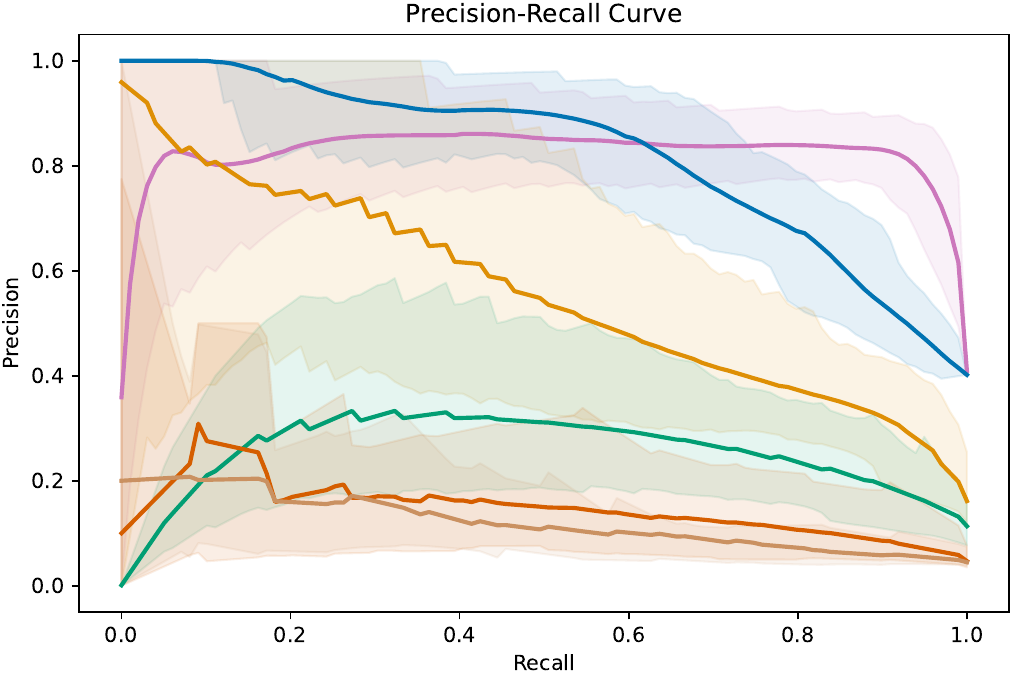}
    \includegraphics[width=0.98\textwidth]{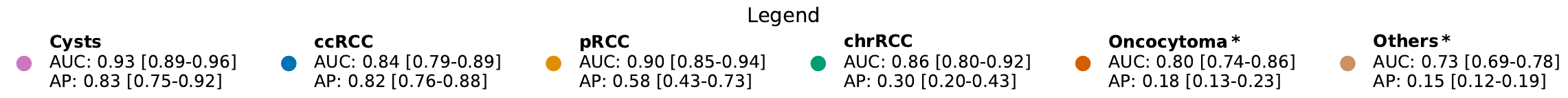}
    \caption{Performance of the radiomics model. The figures show class-wise AUC and AP, alongside their corresponding 95\% confidence intervals, with a combined legend. Cysts, clear cell renal cell carcinoma (ccRCC), papillary RCC, and chromophobe RCC were evaluated on external TCIA data; *Oncocytoma and "Others" through 10-fold cross-validation on internal validation data.}
\else
    \refstepcounter{figure}
\fi
\label{fig:results_radiomics}
\end{figure}

To qualitatively assess the learned representations, we visualized the feature spaces of the training data using UMAP (Fig. \ref{fig:UMAP_embeddings}). Across all embeddings, cysts and ccRCC demonstrated clear separation, most prominently in the CTFM feature space. In contrast, the remaining classes showed substantial overlap and did not form distinct clusters.
Class-wise performance was evaluated using the radiomics model, which achieved the strongest overall class-level results (Fig.~\ref{fig:results_radiomics}). On the external test set, discrimination of cysts (AUC 0.93, AP 0.83) and ccRCC (AUC 0.84, AP 0.82) was robust. Although pRCC and chRCC achieved comparable AUCs (0.90 and 0.86), their AP values were substantially lower (0.58 and 0.30), accompanied by wide confidence intervals, indicating reduced precision and class imbalance effects.
Oncocytoma and “other” lesions were evaluated on the validation set, where performance was lower (Oncocytoma: AUC 0.80, AP 0.18; Others: AUC 0.73, AP 0.15). For comparison, a majority-vote baseline would yield a constant AUC of 0.50 and AP values corresponding to class prevalence (0.55, 0.32, 0.09, 0.05, 0.04, and 0.04 for the respective datasets).
To interpret the XGBoost classifier, features were ranked by relative importance. Grouped by class, texture features contributed the most (0.65), followed by first-order intensity statistics (0.21) and shape (0.14). Top individual predictors were \textit{firstorder\_Energy} (0.059), \textit{glcm\_Autocorrelation} (0.027), and \textit{glcm\_HighGrayLevelRunEmphasis} (0.025). \textit{shape\_SurfaceVolumeRatio} (0.017) was the most influential shape feature, ranking 8th of 107.

\begin{table}[htbp]
\ifPrintTables
    \caption{Performance metrics for Internal Validation (6 classes) and External Test (4 classes). Test scores are reported as Mean with 95\% Confidence Intervals (CI) in parentheses. Best results are marked in bold.}
    \label{tab:combined_results}
    \centering
    \small 
    
    \newcommand{\res}[2]{%
        \makecell[c]{#1 \\[-0.5ex] \footnotesize \color{gray}(#2)}%
    }
    \newcommand{\resbold}[2]{%
        \makecell[c]{\textbf{#1} \\[-0.5ex] \footnotesize \color{gray}(\textbf{#2})}%
    }
    \newcommand{\meanonly}[1]{%
        \makecell[c]{#1}%
    }
    \begin{NiceTabular}{ll cc cc}[colortbl-like]
    \CodeBefore
    \rowcolor{gray!20}{1,2}
    \Body
    \toprule
    \textbf{Backbone} & \textbf{Method} & \multicolumn{2}{c}{\textbf{Int. Val. (6 cls)}} & \multicolumn{3}{c}{\textbf{ Ext. Test (4 cls)}} \\
    \cmidrule(lr){3-4} \cmidrule(lr){5-7}
     & & \textbf{AUC} & \textbf{AP} & \textbf{AUC} & \textbf{AP}\\
    \midrule
    Maj. Vote & Guessing & \meanonly{0.50} & \meanonly{0.25} & \meanonly{0.50} & \meanonly{0.17} \\
    \midrule
    ResNet-50 & Scratch & \res{0.64}{0.62--0.66} & \res{0.28}{0.27--0.29} & \res{0.72}{0.68--0.77} & \res{0.45}{0.40--0.51} \\
    Radiomics    & XGBoost & \resbold{0.84}{0.82--0.86} & \resbold{0.45}{0.42--0.48} & \resbold{0.88}{0.85--0.91} & \resbold{0.64}{0.58--0.70} \\
    \midrule
    CT-FM        & XGBoost & \res{0.67}{0.64--0.71} & \res{0.31}{0.29--0.32} & \res{0.70}{0.65--0.74} & \res{0.43}{0.38--0.49} \\
    MMM          & XGBoost & \res{0.75}{0.72--0.77} & \res{0.36}{0.34--0.38} & \res{0.77}{0.73--0.82} & \res{0.50}{0.44--0.56} \\
    FMCIB        & XGBoost & \res{0.71}{0.70--0.73} & \res{0.33}{0.31--0.34} & \res{0.69}{0.64--0.74} & \res{0.40}{0.35--0.46} \\
    FMCIB        & fine-tuning  & \res{0.68}{0.65--0.71} & \res{0.32}{0.30--0.34} & \res{0.70}{0.64--0.75} & \res{0.42}{0.37--0.47}  \\
    \bottomrule
    \end{NiceTabular}
\else
    \refstepcounter{table}
    \label{tab:combined_results}
\fi
\end{table}

\begin{table}[htbp]
\ifPrintTables
    \caption{AUC differences between foundation model based classifiers and baseline models on the external TCIA data. Values are denoted as multiple-comparison-adjusted 95\% confidence intervals, accompanied by their respective p-values. Positive values indicate better performance of the baseline, vice versa. Statistical significance (*) was inferred where the 95\% CI excluded zero.
}
    \label{tab:sig}
    \centering
    \small 

    \newcommand{\res}[2]{%
        \makecell[c]{#1 \\[-0.5ex] \footnotesize \color{gray}#2}%
    }
    
    \begin{NiceTabular}{l@{\hskip 20pt}l@{\hskip 20pt}c@{\hskip 20pt}c}[colortbl-like]
    \CodeBefore
    \rowcolor{gray!20}{1}
    \Body
    \toprule
    \textbf{Backbone} & \textbf{Method} & \makecell{\footnotesize \textbf{$\Delta$AUC CI to:}\\\textbf{ResNet-50}} & \makecell{\footnotesize \textbf{$\Delta$AUC CI to:}\\\textbf{Radiomics}} \\
    \midrule
    CT-FM & XGBOOST & \res{\(-0.06\)--\(0.09\)}{p = 0.980} & \res{\(0.10\)--\(0.25\)}{p < 0.001*} \\[2ex] 
    MMM   & XGBOOST & \res{\(-0.13\)--\(0.02\)}{p = 0.307} & \res{\(0.03\)--\(0.18\)}{p = 0.002*} \\[2ex]
    FMCIB & XGBOOST & \res{\(-0.05\)--\(0.10\)}{p = 0.898} & \res{\(0.11\)--\(0.26\)}{p < 0.001*} \\[2ex]
    FMCIB & fine-tuning  & \res{\(-0.06\)--\(0.09\)}{p = 0.968} & \res{\(0.10\)--\(0.26\)}{p < 0.001*} \\
    \bottomrule
    \end{NiceTabular}
\else
    \refstepcounter{table}
    \label{tab:sig}
\fi
\end{table}

\section{Discussion}
Non-invasive, CT-based classification of renal lesions is inherently constrained by the scarcity of clinical data. To determine if medical foundation models (FMs) can mitigate this dependence on large datasets, this study investigated the transfer learning performance of three open-source FMs against a 3D ResNet-50 and a radiomics classifier. Using a frozen feature-probing protocol, we found that the conventional radiomics baseline (AUC 0.88) significantly outperformed all evaluated FM approaches (AUC 0.69--0.77), showing that current medical FM are not yet able to advance renal lesion classification beyond established baselines.

Among deep learning approaches, all four FM configurations performed similarly to the ResNet baseline on external testing (AUC 0.69--0.77 vs.\ 0.72), with no statistical significant differences.
This performance was achieved with a massive reduction in computational cost. While the ResNet baseline required 16 hours of NVIDIA A100 GPU-accelerated training, the MMM probing pipeline only needed 2.5 hours for feature extraction using a 32 core AMD EPYC CPU. Since feature extraction is a one-time operation, independent of the cross-validation folds, the subsequent training of classifiers took only 56 seconds per fold. This efficiency makes the approach much more practical for clinical settings that may lack specialized computing resources.
Despite the relative success of FMs, the radiomics baseline significantly outperformed even the best deep learning model (Mean $\Delta$AUC +0.11, 95\% CI [0.03, 0.18]). This suggests that for renal lesion stratification, explicitly defined engineering features may currently capture more discriminative signal than the latent representations of generalist FMs. The interpretability of radiomics, heavily weighted towards texture features, further supports its clinical utility over opaque deep learning embeddings.

FMs are trained on larger anatomical contexts and may not optimize for intra-lesion texture. Nevertheless, all pre-trained FMs were able to extract meaningful features from the CT scans, as can be seen by the clear separation between ccRCC and Cysts in the low-dimensional UMAP visualizations. This qualitative behaviour aligns with our quantitative results where these distinct subtypes had higher classification metrics compared to the more challenging pRCC and chrRCC classes. 
While prior work \supercite{pai2024foundation,schafer2024overcoming} suggests that fine-tuning FMs yields additional gains, our fine-tuned FMCIB model did not significantly differ from the from-scratch ResNet. Although pre-training accelerated training convergence, it did not improve external generalization.
This highlights a broader challenge: comparisons between the individual FMs, fine-tuned or used as frozen encoders, must be interpreted with caution. Unlike established radiomics protocols, there is currently no consensus on the optimal downstream utilization of medical FMs. Critical implementation details, such as patch size, resampling resolution, normalization, and feature aggregation methods, are often under-documented and left to user discretion. This ambiguity complicates the isolation of intrinsic architectural advantages from hyperparameter sensitivity. Consequently, our study benchmarks the general utility of the FM paradigm against conventional baselines, rather than establishing a definitive ranking of specific FMs. Future research must prioritize standardized "instructions for use" to ensure reproducibility and fair model comparison.

Finally, despite high AUC values, the low Average Precisions across all tested models indicate a false positive rate that currently prohibits autonomous clinical use.
This suggests that the required discriminatory information may be absent from standard CT imaging. Differentiating benign RO from malignant chrRCC, for instance, is inherently difficult due to overlapping enhancement patterns \supercite{choudhary2009renal}, implying that diagnostic accuracy may be bounded by a lack of distinct radiological phenotypes rather than model capacity.
This diagnostic uncertainty might be inflated by selection bias. This study only used CT scans with histopathologically confirmed ground truths. Lesions with clear benign appearance (e.g., angiomyolipoma, simple cysts) are often managed via active surveillance and thus excluded from histopathological datasets. Consequently, our dataset over-represents ambiguous or atypical presentations, making the classification task inherently more difficult than a general screening scenario.

In conclusion, foundation models can be as good as from-scratch deep learning in data-scarce renal lesion stratification while drastically reducing computational cost. However, handcrafted radiomics remained the most accurate approach, indicating that current generalist FM embeddings do not yet capture the fine-grained texture and shape heterogeneity that drives histological subtype discrimination. The observed variability in FM implementation and performance further underscores the need for standardized evaluation protocols. Until the optimization gaps identified here are addressed, radiomics remain a necessary and effective tool for texture-dependent classification tasks.

\section*{Data Availability}
The in-house training data cannot be shared due to internal data governance policies. The KITS dataset is available at: \url{https://github.com/neheller/kits23}. External evaluation data is saved at \url{https://zenodo.org/records/19630298}. The training and inference code, along with the model weights for the radiomics and embedding-extractors, are available at: \url{https://github.com/hhaentze/RenalVision}.

\section*{Acknowledgments}
This work was funded by the European Union (grant no. 101079894). Views and opinions expressed are, however, those of the author(s) only and do not necessarily reflect those of the European Union or European Health and Digital Executive Agency (HADEA). Neither the European Union nor the granting authority can be held responsible for them. The authors acknowledge the Scientific Computing of the IT Division at the Charité - Universitätsmedizin Berlin for providing computational resources that have contributed to the research results reported in this paper. URL: \url{https://www.charite.de/en/research/research_support_services/research_infrastructure/science_it/#c30646061}
The results published here are in whole or part based upon data generated by the TCGA Research Network: \url{http://cancergenome.nih.gov}.

\printbibliography

\end{document}